\documentclass[10pt,twocolumn,letterpaper]{article}

\usepackage{bm}
\usepackage{multirow}
\usepackage{makecell}
\usepackage[pagenumbers]{cvpr} % To force page numbers, e.g. for an arXiv version
\definecolor{cvprblue}{rgb}{0.21,0.49,0.74}
\usepackage[pagebackref,breaklinks,colorlinks,allcolors=cvprblue]{hyperref}

\def\paperID{9232} % *** Enter the Paper ID here
\def\confName{CVPR}
\def\confYear{2025}

\title{HDI-Former: Hybrid Dynamic Interaction ANN-SNN Transformer for Object Detection Using Frames and Events}

\author{Dianze Li\textsuperscript{1,2} \hspace{0.8em} Jianing Li\textsuperscript{1} \hspace{0.8em} Xu Liu\textsuperscript{3,2} \hspace{0.8em} Zhaokun Zhou\textsuperscript{2,1} \hspace{0.8em} Xiaopeng Fan\textsuperscript{3,2} \hspace{0.8em} Yonghong Tian\textsuperscript{1,2}\\ \\
\textsuperscript{1} {\normalsize Peking University}~~~~~~~~~
\textsuperscript{2} {\normalsize Peng Cheng Laboratory}~~~~~~~~~
\textsuperscript{3} {\normalsize Harbin Institute of Technology}
}

\begin{document}
\maketitle
\begin{abstract}
Combining the complementary benefits of frames and events has been widely used for object detection in challenging scenarios. However, most object detection methods use two independent Artificial Neural Network (ANN) branches, limiting cross-modality information interaction across the two visual streams and encountering challenges in extracting temporal cues from event streams with low power consumption. To address these challenges, we propose HDI-Former, a Hybrid Dynamic Interaction ANN-SNN Transformer, marking the first trial to design a directly trained hybrid ANN-SNN architecture for high-accuracy and energy-efficient object detection using frames and events. Technically, we first present a novel semantic-enhanced self-attention mechanism that strengthens the correlation between image encoding tokens within the ANN Transformer branch for better performance. Then, we design a Spiking Swin Transformer branch to model temporal cues from event streams with low power consumption. Finally, we propose a bio-inspired dynamic interaction mechanism between ANN and SNN sub-networks for cross-modality information interaction. The results demonstrate that our HDI-Former outperforms eleven state-of-the-art methods and our four baselines by a large margin. Our SNN branch also shows comparable performance to the ANN with the same architecture while consuming 10.57$\times$ less energy on the DSEC-Detection dataset. Our open-source code is available in the supplementary material.
\end{abstract}    
\section{Introduction}
\label{sec:intro}

DAVIS cameras~\cite{brandli2014240, guo2023three, kodama20231, moeys2017sensitive}, namely hybrid vision sensors, combine a bio-inspired event camera and a conventional frame-based camera within the same pixel array to complement each other. Event cameras~\cite{gallego2020event}, with their high temporal resolution, dynamic range, and low power consumption, detect light changes through dynamic events to overcome conventional camera limitations in high-speed motion blur and challenging lighting conditions. Conversely, frame-based cameras capture absolute brightness with fine textures, addressing event camera shortcomings in static or texture-less scenarios. A trend is emerging in leveraging the complementary advantages of frames and events for various computer vision tasks~\cite{gehrig2021combining, li2023sodformer}. Nevertheless, how to design a high-accuracy and energy-efficient object detection model for two heterogeneous visual streams remains a challenging problem.

\begin{figure}[t]
    \centering
    \centerline{\includegraphics[width=\linewidth]{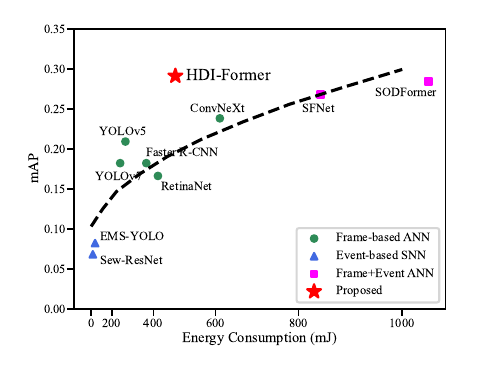}}
    \vspace{-0.1cm}
    \caption{Comparison of our HDI-Former with state-of-the-art methods on the DSEC-Detection dataset. Our HDI-Former achieves a significant mAP improvement over unimodal methods while keeping comparable energy consumption, and it outperforms multimodal methods while reducing energy consumption.}
    \label{fig:scatter}
    \vspace{-0.50cm}
\end{figure}

Most multimodal object detectors rely on two separate branches of Artificial Neural Networks (ANNs)~\cite{cao2021fusion, cao2023chasing, chen2018pseudo, jiang2019mixed, jiang2023nighttime, li2023object, li2023sodformer, li2019event, liu2021attention, liu2023enhancing, luo2023transcodnet, tomy2022fusing, zhou2023rgb}, limiting the cross-modality interaction between two visual streams and resulting in high energy consumption. More specifically, most existing methods typically convert asynchronous events into an image-like representation~\cite{gehrig2019end, vemprala2021representation} and then directly input them into the ANN detection backbone (e.g., SSD~\cite{liu2016ssd} and YOLO~\cite{redmon2018yolov3}). However, Convolutional Neural Networks (CNNs) have limited global spatial modeling capabilities, resulting in lower performance compared to Transformers~\cite{khan2022transformers}, and processing event streams with ANNs involves high computational complexity and energy consumption. Additionally, two separate ANN branches process images and events individually and then generate the final result using a fusion module~\cite{cao2021fusion, liu2021attention, li2023sodformer} in this detection paradigm. Nevertheless, processing two visual streams independently ignores cross-modality information interaction which is crucial for the complementary utilization of fine texture frames and dynamic events. In fact, the biological visual system processes fine texture and dynamic sensing in a more interconnected manner than previously thought~\cite{kang2021retinomorphic, stewart2020review}. Yet, there is no work to design a dynamic interaction neural network with lower power consumption specifically for object detection.

Spiking Neural Networks (SNNs)~\cite{deng2020rethinking, li2024seenn, roy2019towards, zhu2022event} excel in capturing temporal dynamics for event-based vision with low power consumption, but they are rarely used for multimodal object detection tasks. In general, SNNs utilize binary spikes for neural communication, potentially enabling energy-efficient computation~\cite{han2020deep, maass1997networks}. Most SNN object detectors~\cite{chakraborty2021fully, kim2020spiking, hu2023fast, li2022spike} directly convert ANNs to SNNs, relying on large timesteps and the performance of the original ANN. Recent advancements~\cite{bodden2024spiking, cordone2022object, fan2024sfod, su2023deep, yuan2024trainable} using surrogate gradients~\cite{neftci2019surrogate} allow for direct training of SNNs, achieving energy-efficient object detection with fewer timesteps. Nonetheless, these single-modality SNNs, processing only event streams, may be hard to achieve high-accuracy detection. A promising method~\cite{zhao2022framework} is to design hybrid neural networks by combining ANNs and SNNs to leverage the strengths of both. Although some hybrid neural networks~\cite{aydin2023hybrid, yang2019dashnet} integrate ANNs and SNNs to process two heterogeneous visual streams, they are not explicitly designed for object detection. Furthermore, their sub-networks operate independently and are not directly trained, which may hinder them from fully leveraging the complementary benefits of frames and events.

To address the aforementioned problems, we propose a Hybrid Dynamic Interaction ANN-SNN Transformer, namely HDI-Former, which is the first to directly train a hybrid ANN-SNN architecture for high-accuracy and energy-efficient object detection using frames and events. Indeed, this work does not optimize two independent CNN-based branches. In contrast, it seeks to overcome the following challenges: (i) How can we design a novel attention mechanism to achieve high accuracy in the ANN Transformer branch? (ii) How do we create an energy-efficient SNN Transformer branch for event-based object detection? (iii) What is the interaction mechanism for hybrid ANN-SNN branches that benefits two streams? To this end, we first present a novel semantic-enhanced self-attention mechanism to enhance the correlation between image encoding tokens within the ANN Transformer branch. Then, we design an energy-efficient Spiking Swin Transformer branch to use rich temporal cues from event streams. Finally, we introduce a bio-inspired dynamic interaction mechanism between ANN and SNN sub-networks to achieve cross-modality interaction. The results show that our HDI-Former outperforms the state-of-the-art methods by a large margin and our SNN branch achieves comparable performance to the corresponding ANN with less energy consumption. As shown in Fig.~\ref{fig:scatter}, our HDI-Former reaches the best balance between performance and energy consumption. The main contributions of this paper are summarized as follows:

\begin{itemize}
\item We propose a directly trained Hybrid Dynamic Interaction ANN-SNN Transformer (i.e., HDI-Former), which achieves high-accuracy and energy-efficient object detection using frames and events.
\item We present a novel semantic-enhanced self-attention in the ANN branch to enhance the correlation between image encoding tokens for better detection performance.
\item We design an energy-efficient Spiking Swin Transformer branch that achieves the best performance in directly-trained SNNs for event-based data while matching ANN performance with 0.1$\times$ energy consumption.

% reducing 10.57x and 6.34x less energy on the DSEC-Detection dataset and the Gen1 dataset, respectively.
\item We introduce a bio-inspired dynamic interaction mechanism between ANN and SNN sub-networks to enhance cross-modality information interaction for leveraging the complementarity of frames and events.
\end{itemize}

\section{Related Work}
\label{sec:related_work}

\begin{figure*}[htbp]
    \centering
    \centerline{\includegraphics[width=\linewidth]{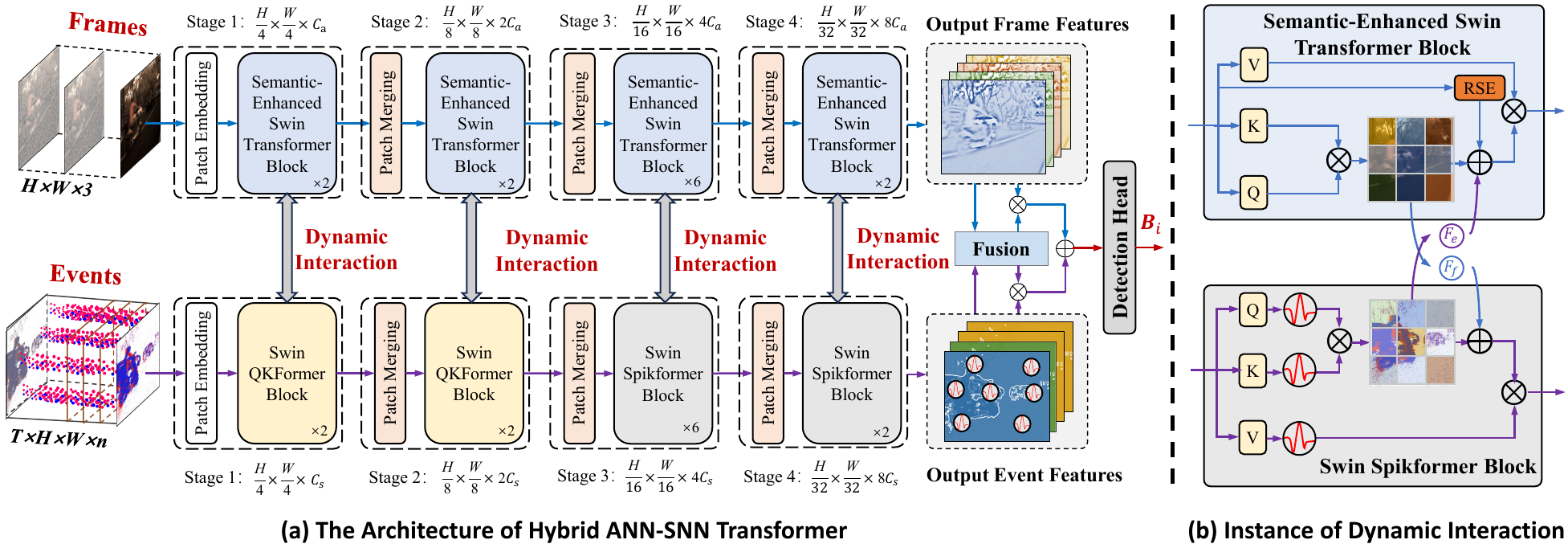}}
    \caption{\textbf{The pipeline of Hybrid Dynamic Interaction ANN-SNN Transformer (HDI-Former)}. (a) Our HDI-Former consists of two sub-networks and connects an ANN and an SNN via dynamic interaction. The ANN branch processes frames using Semantic-Enhanced Swin Transformer blocks, while the SNN branch handles events with Swin QKFormer and Swin Spikformer blocks. Finally, an asynchronous fusion module~\cite{li2023sodformer} combines two streams to predict the detection results. (b) A bio-inspired Dynamic Interaction mechanism between ANN and SNN sub-networks to achieve cross-modality interaction and leverage the complementarity of two streams.}
    \label{fig:framework}
    \vspace{-0.50cm}
\end{figure*}

\textbf{Neuromorphic Object Detection.}
The existing object detectors~\cite{hu2020learning, li2023sodformer} utilizing event cameras can be broadly classified into two categories. The first category~\cite{iacono2018towards, li2022asynchronous, peng2023better, perot2020learning, wang2023dual, zubic2023chaos} involves single-modality approaches that solely process the event stream. These methods typically convert events into 2D image-like representations, which are then fed into feed-forward CNN-based backbones~\cite{iacono2018towards, wang2023dual, zubic2023chaos} or RNN-based object detectors~\cite{li2022asynchronous, peng2023better, perot2020learning}. While these ANNs exhibit high performance, they also come with high computational complexity and energy consumption. More recently, there has been exploration into energy-efficient SNN-based object detectors, including ANN-to-SNN conversion models~\cite{chakraborty2021fully, kim2020spiking, hu2023fast, li2022spike} and directly trained SNNs~\cite{bodden2024spiking, cordone2022object, fan2024sfod, su2023deep, yuan2024trainable}. However, it's evident that relying solely on dynamic event streams may not achieve satisfactory performance in static or slow-motion scenarios that require precise detection of static fine textures. The second category comprises multimodal approaches that combine frames and events. Existing multimodal object detectors~\cite{jiang2019mixed, jiang2023nighttime, li2023sodformer, li2019event, luo2023transcodnet} utilize two independent CNN branches to handle each visual stream and then predict joint detection results using post-processing or end-to-end aggregation fusion modules. However, one drawback is that the limited spatial modeling ability of these CNN models results in a significant performance gap compared to mainstream Transformer models~\cite{gehrig2023recurrent, peng2023get}. Another issue is that two independent ANN branches processing heterogeneous visual streams in the same manner would neglect cross-modality interaction, leading to incomplete utilization of the complementary attributes from frames and events. Thus, this work aims to design a hybrid Transformer with low power consumption to fully leverage the complementarity of two heterogeneous visual streams.

\noindent \textbf{Hybrid Neural Networks.}
Hybrid neural networks~\cite{zhao2022framework} aim to achieve both high accuracy and energy efficiency by combining the strengths of ANNs and SNNs. Due to the development of hybrid vision sensors~\cite{brandli2014240, guo2023three, kodama20231, moeys2017sensitive} and neuromorphic computing hardware, there is a growing trend to design hybrid neural networks. In one class of methods~\cite{kugele2021hybrid, lee2020spike, liu2022enhancing, ahmed2024hybrid}, SNNs are employed for efficient encoding and feature extraction of spatiotemporal events, followed by different ANN backbones for various visual tasks. However, these single-modality hybrid neural networks are only suitable for processing mere event streams. Another type of methods~\cite{aydin2023hybrid, yang2019dashnet, zhao2022framework} adopts two branches of ANNs and SNNs to process multimodal frame and event streams simultaneously. The ANN is tasked with extracting features from the frames while the SNN processes events directly. For example, Aydin et al.~\cite{aydin2023hybrid} design an energy-efficient hybrid ANN-SNN architecture for human pose estimation. Zhao et al.~\cite{zhao2022framework} propose the general design and computation of hybrid neural networks by hybrid units for object tracking. Nevertheless, these hybrid models have not explored the object detection task. Meanwhile, the backbones of their two branches are usually independent of each other, making it difficult for them to interact with each other. Therefore, we propose a dynamic interaction hybrid ANN-SNN model to process two heterogeneous visual streams.

\section{Method}

\subsection{Dynamic Interaction ANN-SNN Framework}
\label{met:dif}
This work aims to design a directly-trained Hybrid Dynamic Interaction ANN-SNN Transformer (HDI-Former), which makes complementary use of frames and events to maximize object detection performance. As shown in Fig.~\ref{fig:framework}, our HDI-Former consists of three novel modules: Semantic-Enhanced Swin Transformer for processing RGB frames (ANN), Spiking Swin Transformer for modeling event streams (SNN), and Dynamic Interaction between them. More precisely, we adopt the Swin Transformer~\cite{liu2021swin} as the backbone since it achieves state-of-the-art performance among Transformer architectures. We then enhance the correlation of tokens by involving relative semantic information to further extract fine-grained spatial features from RGB frames. Then, we present a Spiking Swin Transformer to leverage the rich temporal cues from the continuous event stream. Meanwhile, a bio-inspired dynamic interaction between ANN and SNN sub-networks is designed to facilitate cross-modality information interaction. Finally, an asynchronous fusion module~\cite{li2023sodformer} is utilized to combine two visual flow features complementarily, feeding them into the detection head to predict the final results.

Aiming to implement cross-modality interaction for better utilization of the complementarity of frames and events, we propose a bio-inspired dynamic interaction mechanism between ANN and SNN sub-networks as motivated by the biological vision system in perceiving dynamic and fine textures~\cite{nuthmann2022visual, stewart2020review}. Specifically, this operation makes complementary use of the two modalities and integrates spatiotemporal representations by referencing each other's attention maps. To illustrate our dynamic interaction, we consider an arbitrary block (i.e., $B_l, l=1,2,\cdots$) in both branches. We first compute self-attention in block $B_{i}$ of the SNN branch and the ANN branch to obtain attention weight maps, which can be summarized as follows:
\begin{eqnarray}
    \label{eqn:1}
    W^{e}_{l} = {\rm SoftMax}(\frac{Q_eK_e^\top}{\sqrt{d_e}}+\lambda_1\mathcal{P}+\lambda_2\mathcal{S}_e), \\
    \label{eqn:2}
    W^{f}_{l} = {\rm SoftMax}(\frac{Q_fK_f^\top}{\sqrt{d_f}}+\lambda_1\mathcal{P}+\lambda_2\mathcal{S}_f),
\end{eqnarray}
where $W^{e}_{l},W^{f}_{l} \in \mathbb{R}^{nW\times M^2\times M^2}$ are the attention weight maps for events and frames. $Q_e, Q_f$ and $K_e, K_f$ are queries and keys for the two modalities. $\mathcal{P}$ refers to the relative positional embedding, and $\mathcal{S}_e, \mathcal{S}_f$ are relative semantic embeddings (see Sec.\ref{met:sest}). $nW$ stands for the number of windows, and $M$ is the window size. $\lambda_1,\lambda_2$ are two scalars to reweight $\mathcal{P}$ and $\mathcal{S}$ before adding to the attention weights.

In practice, the attention weight map from each modality is passed to the other and mapped to a unifying attention space using specially designed functions. This projection is necessary because, with multi-head attention, the weights resemble normal feature maps, treating the head dimension like a channel dimension and the attention weight maps for different modalities are distributed over different attention spaces. We refer to this meticulously handcrafted function as the attention kernel function, denoted as $\bm{\mathcal{F}}$. A simple yet effective implementation of the attention kernel function can be mathematically described as:
\begin{equation}
\label{eqn:3}
    \bm{\mathcal{F}}(\bm{A}_{ij})=\frac{1}{H}\sum_{h=1}^H \bm{A}_{ij}^h,
\end{equation}
where $H$ denotes the head number in multi-head self-attention, and $\bm{A} \in \mathbb{R}^{nW\times H\times M^2\times M^2}$ is the attention weight map. Finally, we add the mapped attention weights to the attention weights of the other modality to obtain the final attention weights $\mathcal{W}^{e}_{l}$ and $\mathcal{W}^{f}_{l}$ as:
\begin{eqnarray}
    \label{eqn:4}
    \mathcal{W}^{e}_{l} = W^{e}_{l} \oplus \lambda_3\bm{\mathcal{F}}_f(W^{f}_{l}), \\
    \label{eqn:5}
    \mathcal{W}^{f}_{l} = W^{f}_{l} \oplus \lambda_4\bm{\mathcal{F}}_e(W^{e}_{l}),
\end{eqnarray}
where $\oplus$ denotes the element-wise sum operation. We use attention kernel functions $\bm{\mathcal{F}}_f$ and $\bm{\mathcal{F}}_e$ for frames and events respectively due to the disparity of their attention maps.

\subsection{Semantic-Enhanced Swin Transformer}
\label{met:sest}
In HDI-Former, the ANN branch acts as the spatial feature encoder to extract fine static textures from frames. While the Swin Transformer~\cite{liu2021swin} effectively exploits geometric information in images through relative positional embedding, relying solely on this aspect is insufficient~\cite{dosovitskiy2020image, li2022er}. Thus, we introduce Relative Semantic Embedding, referred to as RSE, to improve data utilization and token correlations by explicitly integrating semantic information into self-attention. Concretely, we first obtain the feature map $X \in \mathbb{R}^{N\times C}$ input to self-attention, where $N$ is the sequence length and $C$ is the channel dimension. Then, we compute the relative semantic distance between any two tokens by subtraction, which can be formulated as:
\begin{equation}
\label{eqn:6}
    a_{ij}=X_i-X_j, i,j=1,2,\cdots, N,
\end{equation}
where $a$ $\in$ $\mathbb{R}^{N \times N\times C}$ is the relative semantic distance matrix. Then we convert $a$ to RSE using a Multi-Layer Perceptron (MLP) with a single output channel. Given there are only two semantic relations between tokens in object detection (i.e., belong to the same object or not), the RSE is supposed to be symmetric. Thus, we feed only the upper triangular part of $a$ into the MLP and symmetrically populate the lower triangles with the resulting RSE. The above-mentioned operations can be described as follows:
\begin{equation}
\mathcal{S}_{ij} = \bm{W}_{m} (\bm{W}_{m}^{'}a_{ij} + \bm{B}_{m}^{'}) + \bm{B}_{m},  i\leq j,
\end{equation}
where $\mathcal{S}\in \mathbb{R}^{N\times N\times1}$ denotes the proposed RSE. $\bm{W}_m, \bm{W}_m'$ and $\bm{B}_m, \bm{B}_m'$ are learnable weights.

Upon RSE, we introduce a novel self-attention mechanism called Semantic-Enhanced Multi-head Self-Attention (SEMSA). By replacing the original self-attention in the Swin Transformer with SEMSA, we create the Semantic-Enhanced Swin Transformer (SEST) as the ANN branch of our HDI-Former. SEST comprises multiple stacked blocks, each executing shifted window partitioning, SEMSA, and feed-forward (FFN) operations. Besides, layer normalization~\cite{xu2019understanding} and dropout~\cite{srivastava2014dropout} are also applied within each block. In summary, two consecutive blocks in our SEST can be formulated as follows:
\begin{align}
    &\hat{Y}_l = {\rm W\text{-}SEMSA}({\rm LN}(Y_{l-1})) + Y_{l-1}, \\
    &Y_l = {\rm MLP}({\rm LN}(\hat{Y}_l)) + \hat{Y}_l, \\
    &\hat{Y}_{l+1} = {\rm SW\text{-}SEMSA}({\rm LN}(Y_{l})) + Y_{l}, \\
    &Y_{l+1} = {\rm MLP}({\rm LN}(\hat{Y}_{l+1})) + \hat{Y}_{l+1},
\end{align}
where $\hat{Y}_l, \hat{Y}_{l+1}\in \mathbb{R}^{N\times C}$ denote the outputs of W-SEMSA and SW-SEMSA modules. $Y_l$ and $Y_{l+1}$ are the outputs of the two blocks. W-SEMSA and SW-SEMSA represent window-based SEMSA with regular and shifted window partitioning configurations, respectively.

\subsection{Spiking Swin Transformer}

\begin{figure}[t]
  \centering
   \includegraphics[width=\linewidth]{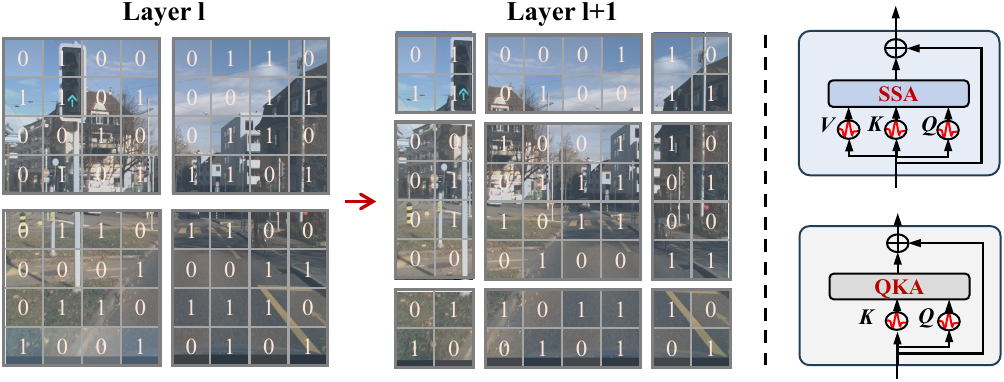}
   \caption{The spiked self-attention with shifted windows and two kinds of blocks in the Spiking Swin Transformer.}
   \label{fig:spiking_swin}
   \vspace{-0.5cm}
\end{figure}

To efficiently process event streams with low energy consumption and enable dynamic interaction with the ANN branch, we design a novel spiking version of the Swin Transformer, which is the first trial to design a directly-trained SNN-based Transformer model for energy-efficient object detection. In contrast to ANN, SNN neurons transmit binary spikes for communication~\cite{zhou2022spikformer}, so we need to convert the feature map $I$ into binary spikes and make the core self-attention mechanism compatible with spikes. To achieve this, we generate the binary feature map by adding a spiking neuron layer after the first convolutional layer in the patch splitting stage. Moreover, the Q, K, V produced by self-attention are quantized by binarization (see Fig.~\ref{fig:spiking_swin}). Meanwhile, we introduce the Spiking Self-Attention (SSA) to extract the spiked feature maps. By removing softmax normalization, SSA completely avoids multiplications, aligning with the energy-efficient calculations of SNNs. The operational steps in our Spiking Swin Transformer can be mathematically represented as follows:
\begin{align}
    & {\rm SSA'}(X) = \mathcal{SN}((QK^\top\cdot s+\lambda_1\mathcal{P}+\lambda_2\mathcal{S}_e)V), \\
    & {\rm SSA}(X) = \mathcal{SN}({\rm BN}({\rm Linear}({\rm SSA'}(X)))), \\
    & \hat{Z}_l = {\rm SSA}(Z_{l-1}) + Z_{l-1}, \\
    & Z_l = {\rm MLP}(\hat{Z}_l) + \hat{Z}_l,
\end{align}
where $Q, K, V$ refer to spiked query, key, and value derived from $X$. $\mathcal{SN}$ represents spiking neuron layer, and $s$ is a scaling factor. $\hat{Z}_l$ and $Z_l$ are the output of SSA and the block. Batch normalization is implemented for re-parameterization convolutions as~\cite{ding2021repvgg} suggests.

Due to the need for a higher frequency of processing events compared to frames during asynchronous inference and the challenges in training SNNs, we propose a new training strategy for object detection regression tasks. In short, we introduce Q-K attention to make our Spiking Swin Transformer more efficient and easier to train. Q-K attention only uses two spike-form components query and key. By performing row summation, Q-K attention derives a token-wise attention vector $A_t\in \mathbb{R}^{N\times 1}$ from $Q$, which models the binary importance of different tokens. Then, Hadamard product $\otimes$ is conducted between $A_t$ and $K$. We can formulate these operations as follows:
\begin{align}
   &  A_t = {\rm \mathcal{SN}}(\sum_{j=1}^D Q_{ij}), \\
   & {\rm QKA}(\bm{X}) = A_t \otimes K,
\end{align}
where ${\rm QKA}$ stands for the Q-K attention operation. In real implementation, we utilize both QKA and SSA for a trade-off between speed and accuracy.

\section{Experiments}
\label{sec:experiment}

\subsection{Experimental Settings}
\label{exp:set}

\begin{table*}[t]
\caption{Comparison with state-of-the-art methods and our baselines* on the DSEC-Detection dataset~\cite{Gehrig24nature}.}
\vspace{-0.2cm}
\footnotesize
\centering
\label{tab:main}
\setlength{\tabcolsep}{2.63 mm}{
\begin{tabular}{c|ccccccc}
\toprule
\textbf{Modality} & \textbf{Method} & \textbf{Type} & \textbf{Backbone} & \textbf{Head} & \textbf{\#Params} & \textbf{mAP$_{50}$} & \textbf{mAP} \\ \hline
\multicolumn{1}{c|}{\multirow{8}{*}{RGB}} & Faster R-CNN~\cite{ren2015faster} & ANN & ResNet50+FPN & Faster R-CNN & 41.9M & 0.354 & 0.182 \\
 & RetinaNet~\cite{lin2017focal} & ANN & ResNeXt-101-FPN & RetinaHead & 53.9M & 0.305 & 0.166 \\
 & CenterNet~\cite{zhou2019objects} & ANN & Hourglass-104 & CenterNet & 191.4M & 0.351 & 0.204 \\
 & YOLOv5~\cite{zhu2021tph} & ANN & CSPDarknet53 & YOLOv5 & 76.8M & 0.332 & 0.209 \\
 & YOLOv7~\cite{wang2023yolov7} & ANN & E-ELAN & YOLOv7 & 151.7M & 0.315 & 0.182 \\
 & ConvNeXt~\cite{liu2022convnet} & ANN & ConvNeXt-T & Mask R-CNN & 45.6M & 0.462 & 0.248 \\
 & Swin Transformer~\cite{liu2021swin} & ANN & Swin-T & Mask R-CNN & 44.8M & 0.487 & 0.269\\
 & \textbf{SEST}* & ANN & Swin-T & Mask R-CNN & 44.8M & \textbf{0.504} & \textbf{0.277}\\ \hline

\multicolumn{1}{c|}{\multirow{3}*{Event}} & EMS-YOLO~\cite{su2023deep} & SNN & EMS-ResNet10 & YOLOv3 & 6.2M & 0.202 & 0.082 \\
 & Spiking Swin Transformer* & SNN & Swin QKFormer & YOLOv3 & 20.3M & 0.254 & 0.116 \\ 
 & Swin Transformer* & ANN & Swin-T & YOLOv3 & 44.8M & 0.266 & 0.133\\ \hline

\multicolumn{1}{c|}{\multirow{5}{*}{RGB +Events}} & FPN-fusion~\cite{tomy2022fusing} & ANN & ResNet-50 & RetinaHead & / & 0.289 & 0.152 \\
 & SFNet~\cite{liu2023enhancing} & ANN & CSPDarknet53 & YOLOv5 & / & 0.412 & 0.268 \\
 & HDI-Former* & Hybrid & Swin-T + Swin QKFormer & Mask R-CNN & 64.2M & 0.511 & 0.284 \\
 & SODFormer~\cite{li2023sodformer} & ANN & Deformable DETR & Deformable DETR & 84.8M & 0.513 & 0.284 \\
 & \textbf{HDI-Former} & Hybrid & Swin-T + Swin QKFormer & Mask R-CNN & 64.2M & \textbf{0.522} & \textbf{0.291} \\ \bottomrule
\end{tabular}}
\end{table*}

\begin{figure*}[t]
    \centering
    \centerline{\includegraphics[width=\linewidth]{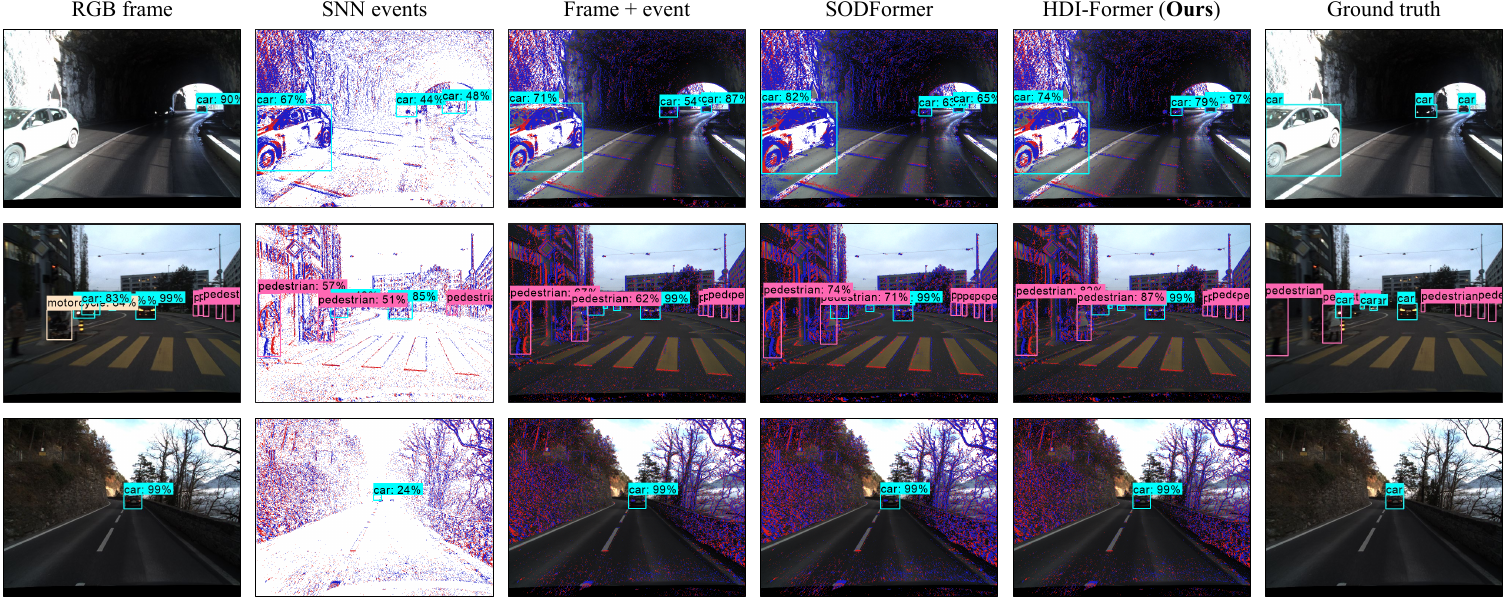}}
    \caption{Representative examples of different object detection results in various scenarios on the DSEC-Detection dataset~\cite{Gehrig24nature}.}
    \label{fig:main_instance}
    \vspace{-0.5cm}
\end{figure*}

\textbf{Dataset.} To verify the effectiveness of our HDI-Former, we conduct experiments primarily on the DSEC-Detection dataset~\cite{Gehrig24nature}. It provides paired frames and events at a resolution of 640$\times$480, along with 390k boxes at 20 Hz for object detection in driving scenarios. For fair comparisons, we also evaluate our Spiking Swin Transformer on the Gen1 dataset~\cite{perot2020learning} which is widely used in SNNs for single-modality event-based object detection. The Gen1 dataset contains events at a resolution of 304$\times$240 and 255k boxes with a frequency of 1 Hz, 2 Hz, or 4 Hz.

\noindent\textbf{Implementation Details.} We select the Swin Transformer-Tiny~\cite{liu2021swin} (i.e., Swin-T) as the backbone owning to accuracy-speed trade-off. The patch size and window size are set to $P$ = 4 and $M$ = 8. In the ANN branch, we set the channel number of the hidden layers in the first stage $C_a$ = 96, while the query dimension of each head is $d_a$ = 32. For the convenience of training SNN, we reduce the number of channels to $C_s$ = 64 and $d_s$ = 16. The timestep in SNN is set to $T$ = 5, and the number of blocks to perform dynamic interaction is $N_l$ = 4. Following Swin Transformer, we use Mask R-CNN~\cite{he2017mask} as our detection head and utilize random resize as data augmentation. With batch size set to 16, we adopt an AdamW optimizer~\cite{kingma2014adam} with an initial learning rate of $10^{-4}$ and weight decay of 0.05. All experiments are conducted on NVIDIA A100 GPUs. We report the mean average precision (e.g., COCO mAP~\cite{lin2014microsoft}) as evaluation metrics, containing mAP at IoU = 0.5 (i.e., mAP$_{50}$) and the average AP between 0.5 and 0.95 (i.e., mAP). We also compare the SNNs from the perspective of energy consumption. %More information about our implementation details can be found in the Appendix.

\subsection{Comparison with State-of-the-Art Methods}
\label{exp:sota}

To verify the effectiveness of HDI-Former, we compare it with eleven state-of-the-art methods and four baselines in Table~\ref{tab:main} and Fig.~\ref{fig:main_instance} to show the advantages of our hybrid framework from three perspectives. 

\noindent\textbf{Frame Modality.} We present a comparison between our SEST branch and seven state-of-the-art methods, including Faster R-CNN~\cite{ren2015faster}, RetinaNet~\cite{lin2017focal}, CenterNet~\cite{zhou2019objects}, YOLOv5~\cite{zhu2021tph}, YOLOv7~\cite{wang2023yolov7}, ConvNeXt~\cite{liu2022convnet}, and Swin Transformer~\cite{liu2021swin}. Note that, our SEST outperforms the former six frame-based feedforward methods by a large margin. This is caused by the fact that the self-attention operation is better at extracting global features compared to standard convolution, indicating the potential of Transformer in the object detection task. Furthermore, by explicitly leveraging both geometric and semantic information, our SEST module comprehensively enhances the correlations between image tokens. Table~\ref{tab:main} shows that our SEST achieves a remarkable mAP improvement of 0.8\% with comparable parameters over the Swin Transformer.

\noindent\textbf{Event Modality.} We compare our Spiking Swin Transformer in the SNN branch with EMS-YOLO~\cite{su2023deep} which achieves state-of-the-art performance among directly trained SNNs to the best of our knowledge. From the event modality in Table~\ref{tab:main}, our Spiking Swin Transformer achieves a significant mAP improvement of 3.4\% over EMS-YOLO. In addition, our SNN branch performs relatively comparable to its ANN counterpart while significantly reducing energy consumption (see Sec.\ref{exp:snn}).

\noindent\textbf{Benefit From Multimodal Fusion.} We further compare our HDI-Former with state-of-the-art multimodal methods (i.e., FPN-fusion~\cite{tomy2022fusing}, SFNet~\cite{liu2023enhancing} and SODFormer~\cite{li2023sodformer}) and our baseline (i.e., HDI-Former* without dynamic interaction). Note that our HDI-Former outperforms FPN-fusion and SFNet by a large margin. Moreover, our HDI-Former, using the SNN branch to process the event stream, dramatically reduces energy consumption compared to SODFormer. Furthermore, we present some representative visualization results on the DSEC-Detection dataset~\cite{Gehrig24nature} in Fig.~\ref{fig:main_instance}. The three rows from top to bottom represent low-light, high-speed motion blur and static scenes, respectively. Obviously, our HDI-Former, making complementary use of frames and events, can perform robust object detection in various scenarios. The first row in Fig.~\ref{fig:main_instance} also shows that our HDI-Former, utilizing continuous event streams with rich temporal cues, successfully detects the occluded car on the left, overcoming single-frame object detection challenges.

\subsection{Performance Evaluation of SNNs}
\label{exp:snn}

\begin{figure}[t]
  \centering
   \includegraphics[width=\linewidth]{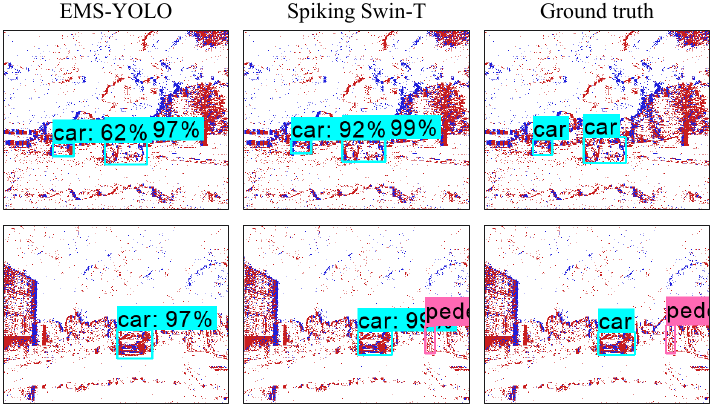}
   \vspace{-0.40cm}
   \caption{Representative visualization results of our SNN-branch and EMS-YOLO on the Gen1 dataset~\cite{de2020large}.}
   \label{fig:gen1_instance}
   \vspace{-0.40cm}
\end{figure}

\textbf{Evaluation on Gen1~\cite{de2020large}.} As illustrated in Table~\ref{tab:Compare SNN on Gen1}, we present a comparison of our Spiking Swin Transformer with some directly-trained SNN-based object detectors on the Gen1 dataset~\cite{de2020large}, the most widely used dataset for single-modality event-based object detection. We can find that our Spiking Swin Transformer achieves the best mAP of 0.291 and mAP$_{50}$ of 0.580 with the same timestep, which outperforms the state-of-the-art methods to our knowledge. It is worth noting that despite the higher performance achieved by the ANN counterpart, our Spiking Swin Transformer exploits a sparse representation with a spike firing rate of 19.84\%, implying lower energy consumption. According to the quantitative formula for energy consumption in~\cite{su2023deep}, we conclude that our Spiking Swin Transformer reduces up to 6.34$\times$ energy consumption compared to its ANN counterpart. Besides, some representative visualization results in Fig.~\ref{fig:gen1_instance} also indicate the effectiveness of our Spiking Swin Transformer on the Gen1 dataset~\cite{de2020large}.

\begin{table}[t]
\caption{Comparison of SNNs on the Gen1 dataset~\cite{de2020large}.}
\vspace{-0.2cm}
\centering
\footnotesize
\label{tab:Compare SNN on Gen1}
\setlength{\tabcolsep}{1.25 mm}{
\begin{tabular}{cccccc}
\toprule
\textbf{Model} & \textbf{Params} & \textbf{Firing Rate} & \textbf{Efficiency} & \textbf{mAP$_{50}$} & \textbf{mAP} \\ \hline
MS-ResNet18    & 9.49M & 17.08\% & 2.43×        & 0.560 & 0.285\\
Sew-ResNet18   & 9.56M & 18.80\% & 2.00×$^{*}$  & 0.561 & 0.286\\
EMS-ResNet18   & 9.34M & 20.09\% & 4.91×        & 0.565 & 0.286 \\
Swin-T (ANN)   & 44.8M & /       & /            & 0.588 & 0.296 \\
Spiking Swin-T & 20.3M & 19.84\% & 6.34×        & 0.580 & 0.291\\ 
\bottomrule
\end{tabular}}
\end{table}

\noindent\textbf{Evaluation on DSEC-Detection~\cite{Gehrig24nature}.} As shown in Table~\ref{tab:Compare SNN on DSEC}, We further explore the effectiveness of our Spiking Swin Transformer on the DSEC-Detection dataset~\cite{Gehrig24nature}, which contains more classes of objects and is relatively more complex. By substituting CNN-based ResNet with the Swin Transformer which is more capable of modeling, our model gains greater expressive ability and outperforms EMS-YOLO~\cite{su2023deep} by a large margin of 3.4\% mAP. Besides, our Spiking Swin Transformer maintains a low firing rate when facing more complicated data while EMS-YOLO increases significantly. As a result, the energy consumption can be saved up to 10.57$\times$ compared to ANN Swin Transformer. In other words, our SNN branch achieves comparable performance to the ANN with the same architecture while significantly reducing power consumption.

\begin{table}[htbp]
\caption{Comparison of SNNs on the DSEC-Detetection dataset.}
\vspace{-0.2cm}
\centering
\footnotesize
\label{tab:Compare SNN on DSEC}
\setlength{\tabcolsep}{2.00mm}{
\begin{tabular}{cccccc}
\toprule
\textbf{Model} & \textbf{Firing Rate} & \textbf{\makecell[c]{Energy\\Consumption}} & \textbf{mAP$_{50}$} & \textbf{mAP} \\ \hline
EMS-ResNet10 & 32.8\%   & 8.18  & 0.202 & 0.082 \\
Swin-T (ANN) & /        & 295.4 & 0.266 & 0.133 \\
Spiking Swin-T & 20.9\% & 27.95 & 0.254 & 0.116\\ \bottomrule
\end{tabular}}
\vspace{-0.1cm}
\end{table}

\begin{table}[t]
\centering
\caption{The contribution of each component to our HDI-Former on the DSEC-Detection dataset~\cite{Gehrig24nature}.}
\vspace{-0.2cm}
\scriptsize
\label{tab:ablation experiment}
\setlength{\tabcolsep}{1.36mm}{
\begin{tabular}{cccccccc}
\toprule
\multirow{3}{*}{\textbf{Method}} & \multirow{3}{*}{\textbf{RSE}} & \multirow{3}{*}{\textbf{Events}} & \multirow{3}{*}{\textbf{\makecell[c]{Dynamic \\ Interaction}}} & \multirow{3}{*}{\textbf{mAP$_{50}$}} & \multirow{3}{*}{\textbf{mAP}} & \multicolumn{2}{c}{\textbf{Ops(G)}} \\ \cline{7-8} &  &  &  &  & & \multirow{2}{*}{\textbf{AC}} & \multirow{2}{*}{\textbf{MAC}} \\ &  &  &  &  &  &  \\ \hline
Baseline &  &  &  & 0.487 & 0.269  & 0  & 98.6  \\
(a) & $\surd $  &  &  & 0.504 & 0.277  & 0  & 99.5  \\
(b) & $\surd $  & $\surd $  &  & 0.511 & 0.284  & 26.12  & 100.17  \\
HDI-Former & $\surd $  & $\surd $  & $\surd $  & \textbf{0.522} & \textbf{0.291} & 26.12  & 100.17 \\ \bottomrule
\end{tabular}}
\vspace{-0.10cm}
\end{table}

\subsection{Ablation Study}
\label{exp:abl}

\textbf{Contribution of Each Component.} We use the ANN Swin Transformer as the baseline to assess the contributions of each module. As illustrated in Table~\ref{tab:ablation experiment}, three methods namely (a), (b) and our HDI-Former consistently achieve higher performance by relying on the proposed components. To be specific, method (a), enhancing the correlations between tokens in self-attention by explicitly utilizing semantic information, obtains a 0.8\% mAP improvement over the baseline. By involving events through the Spiking Swin Transformer, method (b) is capable of exploiting rich temporal cues from continuous event streams and making complementary use of frames and events, so the detection mAP is further enhanced by 0.7\% compared to (a). By adopting dynamic interaction, our HDI-Former achieves an mAP improvement of 0.7\% over method (b) and 1.4\% over method (a). This reveals that our dynamic interaction mechanism favors complementary context capturing across frames and events while improving the model's capability to extract features at a more fundamental level during the feature extraction stage. Besides, our HDI-Former introduces few ACs and almost no MACs compared to the baseline, ensuring high accuracy and energy efficiency.

\begin{table}[t]
\centering
\caption{The influence of SNN timesteps on the Gen1 dataset~\cite{de2020large}.}
\vspace{-0.2cm}
\footnotesize
\label{tab:ablation T}
\setlength{\tabcolsep}{3.60mm}{
\begin{tabular}{cccccc}
\toprule
\textbf{T} & \textbf{1} & \textbf{3} & \textbf{5}     & \textbf{7} & \textbf{9} \\ \hline
mAP$_{50}$ & 0.565      & 0.572      & \textbf{0.580} & 0.568      & 0.564      \\
mAP        & 0.286      & 0.289      & \textbf{0.291} & 0.288      & 0.286      \\ \bottomrule
\end{tabular}}
\vspace{-0.4cm}
\end{table}

\noindent\textbf{Influence of SNN Timesteps.} To explore the impact of SNN timesteps, we report the performance of our Spiking Swin-T on the Gen1 dataset by first fixing the length of temporal window and then setting various timesteps $T$ (see Table~\ref{tab:ablation T}). Note that, the performance is subpar when $T$ is 1 or 3 as the small timestep hinders feature extraction. Conversely, it may be unsatisfactory with a high timestep (e.g., $T$ = 7) due to increased training difficulty. The limited number of events within each timestep may also contribute to the inability to convey valid information when $T$ is too large.

\begin{table}[htbp]
\centering
\caption{The influence of the number of layers conducting dynamic interaction on the DSEC-Detection dataset~\cite{Gehrig24nature}.}
\vspace{-0.2cm}
\footnotesize
\label{tab:ablation dynamic interaction}
\setlength{\tabcolsep}{3.60mm}{
\begin{tabular}{cccccc}
\toprule
\textbf{Layers} & \textbf{0} & \textbf{2} & \textbf{4}     & \textbf{6} & \textbf{8} \\ \hline
mAP$_{50}$      & 0.511      & 0.518      & \textbf{0.522} & 0.521      & 0.517 \\
mAP             & 0.284      & 0.288      & \textbf{0.291} & 0.290      & 0.286 \\ \bottomrule
\end{tabular}}
\vspace{-0.4cm}
\end{table}

\noindent\textbf{Influence of the Number of Interaction Layers.} To further analyze the impact of dynamic interaction, we vary the number of layers where dynamic interaction is applied to 2, 4, 6, and 8, respectively, and evaluate the performance of our HDI-Former. From Table~\ref{tab:ablation dynamic interaction}, it's evident that our HDI-Former achieves optimal performance with 4 dynamic interaction layers ($N_l$ = 4). This suggests that dynamic interaction increases the model's expressiveness. Interestingly, the performance seems to vary marginally across different numbers of dynamic interaction layers, implying that a few layers are sufficient to achieve our objectives.

\subsection{Scalability Test}
\label{exp:sca}

\begin{figure}[htbp]
    \centering
    \centerline{\includegraphics[width=\linewidth]{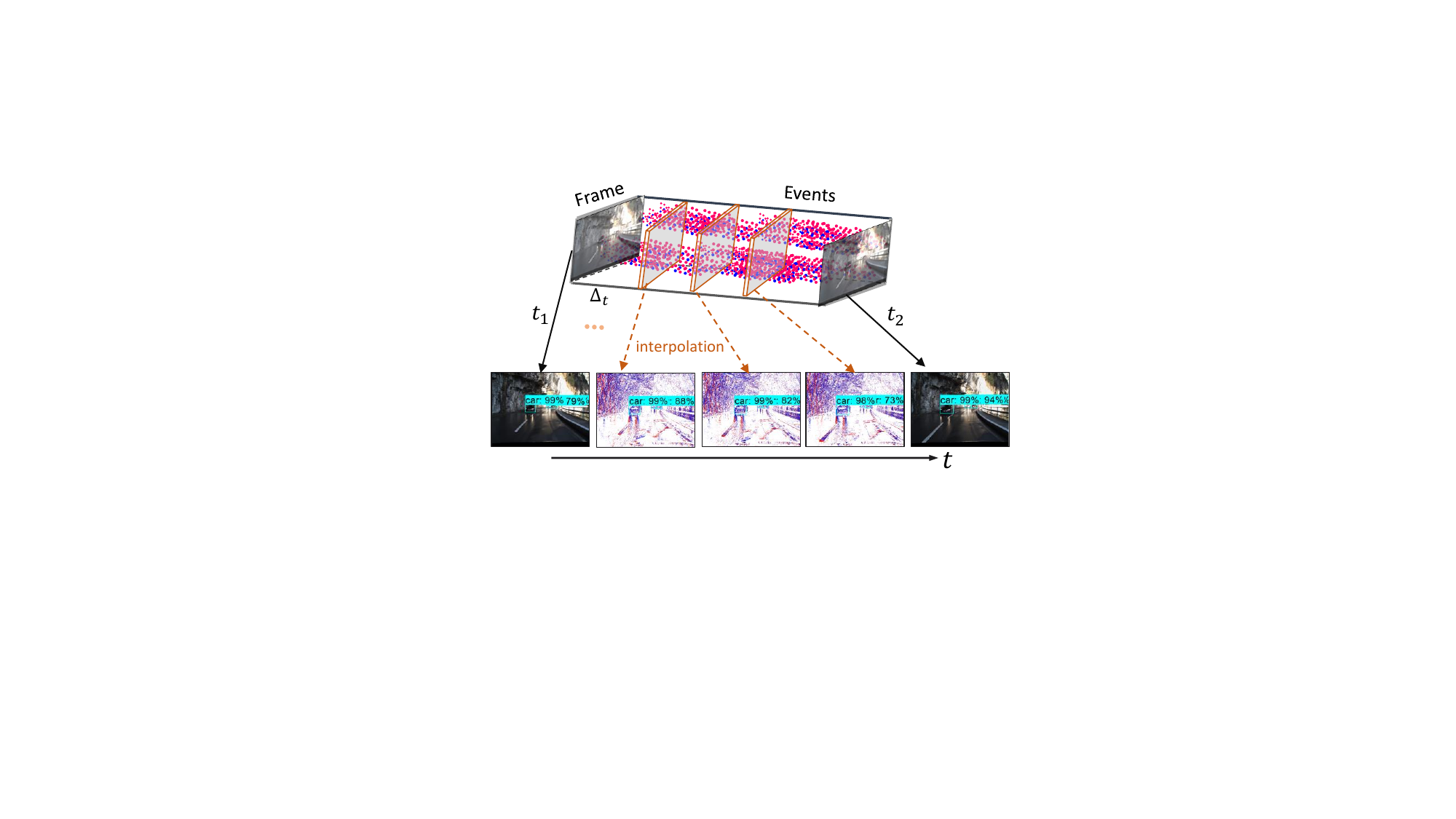}}
    \vspace{-0.1cm}
    \caption{Visualization of asynchronous high-frequency inference through incorporating the continuous event stream.}
    \label{fig:async_infer}
    \vspace{-0.20cm}
\end{figure}

\noindent\textbf{Asynchronous Inference Validation.} Conventional cameras with global shutters face limitations in output frame rate, resulting in gaps between adjacent frames. Our HDI-Former introduces an event stream of high temporal resolution to effectively fill these gaps and achieve continuous object detection. To achieve high-frequency detection, we first sample from the event stream using a sliding window of 0.05s length shifting forward by 0.0125s, resulting in an event stream of 80 Hz. We then input frames at 20 Hz and event temporal bins at 80 Hz into our HDI-Former to produce detections at 80 Hz. As depicted in Fig.~\ref{fig:async_infer}, the middle three figures display the predictions, showing a smooth transition of the predicted box from the first frame to the second. This indicates that our HDI-Former can leverage the event stream of high temporal resolution to overcome the limitation of inference frequency and achieve continuous object detection in real-world applications.

\begin{figure}[htbp]
    \centering
    \centerline{\includegraphics[width=\linewidth]{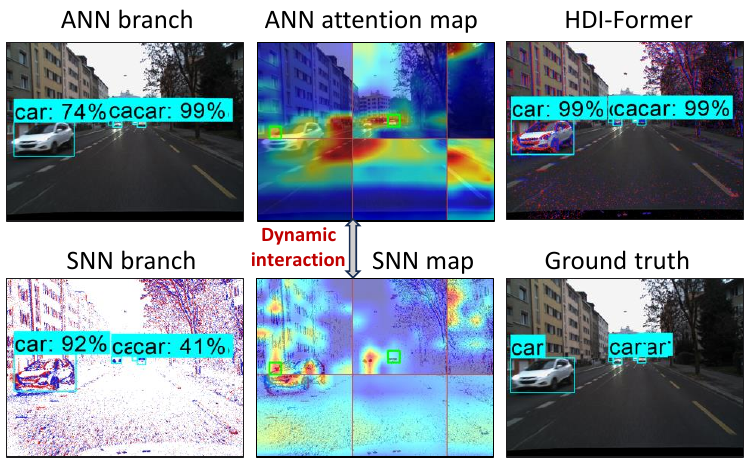}}
    \vspace{-0.2cm}
    \caption{Representative instance verifying the necessity and effectiveness of dynamic interaction.}
    \label{fig:attn_map}
    \vspace{-0.5cm}
\end{figure}

\noindent\textbf{Dynamic Interaction Analysis.} To investigate the interpretability of the dynamic interaction mechanism, we present a scenario involving objects moving at different speeds in Fig.~\ref{fig:attn_map}. We observe a car on the left experiencing high-speed motion blur, while a car in the middle appears static. The attention map reveals that the ANN branch fails to highlight the left object, and the event modality struggles to recognize the middle object. However, these unattended regions contain valuable information crucial for accurate detection. Leveraging attention weights from each other, our dynamic interaction mechanism effectively addresses this challenge. In other words, the proposed dynamic interaction mechanism can leverage the complementarity of frames and events via cross-modality interaction.

\section{Conclusion}

This paper proposes a novel Hybrid Dynamic Interaction ANN-SNN Transformer, namely HDI-Former, which is a pioneering approach to integrate ANNs and SNNs to utilize their complementary strengths dynamically. To achieve this, we first present a novel semantic-enhanced self-attention mechanism to strengthen the correlation between tokens in the ANN branch. Then an energy-efficient Spiking Swin Transformer branch is proposed to extract temporal cues from events. Finally, the dynamic interaction mechanism enables cross-modality interaction between the ANN and SNN. Experimental results demonstrate that our HDI-Former significantly surpasses the state-of-the-art methods and the SNN branch achieves comparable performance to the ANN counterpart at lower energy consumption. To our knowledge, this is the first attempt to directly train a hybrid ANN-SNN Transformer for high-accuracy energy-efficient multimodal object detection. We believe that conducting effective cross-modality interaction is an exciting breakthrough in object detection with frames and events, inspiring directions for future research. 
% To summarize, this study represents an significant progress towards effectively integrating ANNs and SNNs to leverage the complementary advantages of frames and events.

\noindent\textbf{Limitation.} The power consumption is simulated based on existing academic papers rather than neuromorphic chips, potentially increasing inference time and skewing power usage estimates. Future research will focus on deploying algorithms onto neuromorphic chips to meet the high-speed and energy-efficient requirements of event cameras.
{
    \small
    \bibliographystyle{ieeenat_fullname}
    \bibliography{main}
}

\clearpage
\setcounter{page}{1}
\setcounter{section}{0}
\renewcommand\thesection{\Alph{section}}
\renewcommand\thesubsection{\Alph{section}.\arabic{subsection}}
\maketitlesupplementary

\begin{table}[t]
\caption{Comparison with state-of-the-art methods on the DSEC-Detection dataset~\cite{Gehrig24nature} involving only 2 classes.}
\vspace{-0.2cm}
\scriptsize
\centering
\label{tab:twoclass}
\setlength{\tabcolsep}{0.98mm}{
\begin{tabular}{c|ccccc} 
\toprule
\textbf{Modality}            & \textbf{Method}     & \textbf{Type} & \textbf{Backbone} & \textbf{mAP$_{50}$} & \textbf{mAP}   \\ \hline
\multirow{3}{*}{Event}       & RVT~\cite{gehrig2023recurrent}      & ANN           & Transformer+RNN   & 0.587               & 0.384          \\
                             & SAST~\cite{peng2024scene}           & ANN           & Transformer       & 0.601               & 0.381          \\
                             & SSM~\cite{zubic2024state}           & ANN           & Transformer+SSM   & 0.552               & 0.380          \\  \hline
\multirow{4}{*}{Frame+Event} & DAGr-18~\cite{Gehrig24nature}       & ANN           & GNN+CNN           & -                   & 0.376          \\
                             & DAGr-34~\cite{Gehrig24nature}       & ANN           & GNN+CNN           & -                   & 0.390          \\
                             & DAGr-50~\cite{Gehrig24nature}       & ANN           & GNN+CNN           & 0.660               & 0.419          \\
                             & \textbf{HDI-Former} & ANN+SNN       & Transformer       & \textbf{0.691}      & \textbf{0.467} \\
\bottomrule
\end{tabular}}
\end{table}

\section{Detailed DSEC-Detection Dataset}
The DSEC-Detection~\cite{Gehrig24nature} dataset is collected using 2× Prophesee Gen3.1 cameras with a spatial resolution of 640$\times$480 pixels. It includes detection labels for 60 hybrid sequences amounting to 70,379 frames and 390,118 bounding boxes. It is then divided into 47 sequences for training and 13 for testing. The labels are produced by applying an advanced object tracking algorithm and manually reviewing and adjusting to eliminate any incorrect tracks. There exist 8 classes, i.e., pedestrian, rider, car, bus, truck, bicycle, motorcycle, and train. \textbf{\emph{Notably, the results in the main text are all evaluated on the full 8 classes, while some of the methods~\cite{Gehrig24nature} involve only 2 classes (i.e., including only pedestrians and grouping cars, trucks, and buses as cars).}} For comprehensiveness, we further compare HDI-Former with several state-of-the-art methods on the DSEC-Detection dataset with the two-class setup in Table~\ref{tab:twoclass}. The results show that HDI-Former achieves significant performance improvement against the other methods with mAP increases by 4.8\% compared to the second optimal method. Furthermore, HDI-Former utilizes the energy-efficient Spiking Swin Transformer to extract temporal cues from event streams, which reduces energy consumption.

\begin{table*}[t]
\small
\caption{Detailed architecture specifications of the HDI-Former.}
\centering
\label{tab:architecture}
\setlength{\tabcolsep}{6.30mm}{
\begin{tabular}{c|c|c|c|c}
\hline
 & \makecell[c]{Downsp. rate\\(output size)} & Layer name & ANN branch & SNN branch \\ \hline
\multirow{3}{*}{Stage 1} & \multirow{3}{*}{\makecell[c]{4$\times$\\(120$\times$160)}} & Patch Embedding & concat 4$\times$4, 96-d, LN & \makecell[c]{concat 4$\times$4, 64-d,\\BN, $\mathcal{SN}$} \\
\cline{3-5} & & \makecell[c]{SEST /\\ Swin QKFormer} & $\left[ \text{\makecell[c]{win. sz. 8$\times$8,\\dim 96, head 3}} \right]\times 2$ & $\left[ \text{\makecell[c]{win. sz. 8$\times$8,\\dim 64, head 4}} \right]\times 2$ \\ \hline
\multirow{3}{*}{Stage 2} & \multirow{3}{*}{\makecell[c]{8$\times$\\(60$\times$80)}} & Patch Merging & concat 2$\times$2, 192-d, LN & \makecell[c]{concat 2$\times$2, 128-d,\\BN, $\mathcal{SN}$} \\ \cline{3-5} 
 & & \makecell[c]{SEST /\\ Swin QKFormer} & $\left[ \text{\makecell[c]{win. sz. 8$\times$8,\\dim 192, head 6}} \right]\times 2$ & $\left[ \text{\makecell[c]{win. sz. 8$\times$8,\\dim 128, head 8}} \right]\times 2$ \\ \hline
\multirow{3}{*}{Stage 3} & \multirow{3}{*}{\makecell[c]{16$\times$\\(30$\times$40)}} & Patch Merging & concat 2$\times$2, 384-d, LN & \makecell[c]{concat 2$\times$2, 256-d,\\BN, $\mathcal{SN}$} \\ \cline{3-5}
 & & \makecell[c]{SEST /\\ Swin Spikformer} & $\left[ \text{\makecell[c]{win. sz. 8$\times$8,\\dim 384, head 12}} \right]\times 6$ & $\left[ \text{\makecell[c]{win. sz. 8$\times$8,\\dim 256, head 16}} \right]\times 6$ \\ \hline
\multirow{3}{*}{Stage 4} & \multirow{3}{*}{\makecell[c]{32$\times$\\(15$\times$20)}} & Patch Merging & concat 2$\times$2, 768-d, LN & \makecell[c]{concat 2$\times$2, 512-d,\\BN, $\mathcal{SN}$} \\ \cline{3-5} 
 & & \makecell[c]{SEST /\\ Swin Spikformer} & $\left[ \text{\makecell[c]{win. sz. 8$\times$8,\\dim 768, head 24}} \right]\times 2$ & $\left[ \text{\makecell[c]{win. sz. 8$\times$8,\\dim 512, head 32}} \right]\times 2$ \\ \hline
\end{tabular}}
\end{table*}

\section{Preliminaries}

\subsection{Event Camera Sensing Mechanism}

Event cameras, called neuromorphic cameras, report pixel-wise values of $+1$ or $-1$ for positive or negative log intensity differences, respectively.  Each event $\boldsymbol{e_{n}}$ can be described as a four-attribute tuple $(x_{n}, y_{n}, t_{n}, p_{n})$, which is generated for a pixel $\boldsymbol{u} = (x_{n}, y_{n})$ at the timestamp $t_{n}$ once the log-intensity change exceeds the processing threshold $\theta_{th}$, and it can be depicted as:
\begin{eqnarray}
\ln R\left(\boldsymbol{u}_{n}, t_{n}\right)-\ln R\left(\boldsymbol{u}_{n}, t_{n}-\Delta t_{n}\right)=p_{n} \theta_{t h},
\end{eqnarray}
where $\Delta t_{n}$ is the temporal sampling interval at a pixel, and the polarity $p_{n} \in\{-1,1\}$ represents the increasing or decreasing change in the brightness, respectively.

Consequently, asynchronous events $\boldsymbol{S}$=$\left\lbrace  \boldsymbol{e_{n}} \right\rbrace _{n=1}^{N_{e}}$ are sparse and discrete points in the spatio-temporal window $\Gamma$, which can be mathematically described as: 
\begin{equation}
S(x, y, t)=\left\{p_n \delta\left(x-x_n, y-y_n, t-t_n\right)\right\}_{n=1}^{N_e} ,
\end{equation}
where $N_e$ is the number of events during the spatio-temporal window $\Gamma$, and $\delta(\cdot)$ refers to the Dirac delta function, with $\delta(t) = 0, \forall t \not= 0$, and $\int\delta(t) dt = 1$. Event cameras offer the advantages of low latency, high dynamic range, and low redundancy.

\subsection{Theory of Spiking Neural Networks}

In SNNs, spiking neurons are the basic units with more biological plausibility as they mimic the membrane potential dynamics and the spiking communication scheme~\cite{su2023deep}. We choose the Leaky Integrate-and-Fire (LIF) model as the spiking neuron in our work as a good trade-off between bio-plausibility and complexity. The dynamics of a LIF neuron can be formulated as:
\begin{equation}
H(t) = V(t-1) + \frac{1}{\tau}(X(t) - (V(t-1) - V_{reset})) ,
\end{equation}
\begin{equation}
S(t) = \Theta(H(t) - V_{th}) ,
\label{eq: Heaviside step}
\end{equation}
\begin{equation}
V(t) = H(t)(1 - S(t)) +  V_{reset}S(t),
\end{equation}
where $\tau$ is the membrane time constant, and $X(t)$ is the current input at time step $t$. Whenever the membrane potential $H(t)$ exceeds the firing threshold $V_{th}$, the spiking neuron triggers a spike $S(t)$. $\Theta(v)$ is the Heaviside step function, which equals 1 when $v \leq 0$ and 0 otherwise. $V(t)$ represents the membrane potential after the triggered event, which equals $H(t)$ if no spike is generated and otherwise equals the reset potential $V_{reset}$. We can observe that LIF neurons are charged by input stimuli and remain at resting potential at other moments, which accounts for their high efficiency and low energy consumption.

Eq.~\ref{eq: Heaviside step} can be interpreted as a recurrent neural network that can be unrolled over multiple forward Euler steps and then trained using backpropagation through time. However, the Heaviside step function $\Theta(v)$ is not differentiable near 0 which impedes backpropagation. To solve this problem, we use surrogate gradients~\cite{neftci2019surrogate} to replace the gradient of the Heaviside function with the approximate function as:
\begin{equation}
\Theta(v) \approx \frac{1}{1 + (\pi v)^2}.
\end{equation}

SNNs showcase energy-efficient properties with neurons engaging in Accumulation Calculation (AC) operations exclusively during spikes. Hence, the energy consumption of SNNs is a vital metric which can be calculated~\cite{su2023deep} as:
\begin{equation}
E_{SNN}=\sum_{n=1}^{N_b} T_n\left(f_n \cdot E_A \cdot OP_A^n+E_M \cdot OP_M^n\right) ,
\end{equation}
where $T_n$ and $f_n$ are the total time steps and the firing rate in the $n-$th block, respectively. $OP_A^n$ and $OP_M^n$ represent the number of accumulation calculation operations (AC) and Multiply-Accumulate (MAC) operations in the $n-$th block. $E_A$ and $E_M$ denote the computational energy of each operation and directly reference the power consumption of a neuromorphic chip with specific nanometer (nm) technology. In this work, we assume that various operations involved are 32-bit floating-point implemented in 45nm technology, where $E_A$ = 0.9pJ and $E_M$ = 4.6pJ.

\section{Detailed Network Architecture}

\subsection{Network Specifications}
We show more details of the network architecture in Table~\ref{tab:architecture}. It can be seen that we set the architectures of the two branches approximately the same to ensure dynamic interaction. The output sizes are obtained presuming a 640$\times$480 input which is exactly the case in the DSEC-Detection dataset. "concat $n\times n$" represents splitting the feature map into patches of size $n$ and merging the features within each patch into a token, which results in downsampling of feature map by a rate of $n$. In the ANN branch, we achieve this through a Conv2D layer with kernel size and stride both set to $P$ in the patch embedding stage, while patch merging is achieved by concatenating the neighboring $n\times n$ tokens and reducing the channels by 2 with a Linear layer. In the SNN branch, we follow \cite{zhou2024qkformer} to use a \{Conv2D-BN-MaxPooling-$\mathcal{SN}$\} combination for downsampling with a shortcut connection for both patch embedding and patch merging stage. Note that we add an extra \{Conv2D-BN-$\mathcal{SN}$\} layer before the patch embedding stage of the SNN branch to transform the input into spikes.

\begin{figure}[t]
  \centering
   \includegraphics[width=\linewidth]{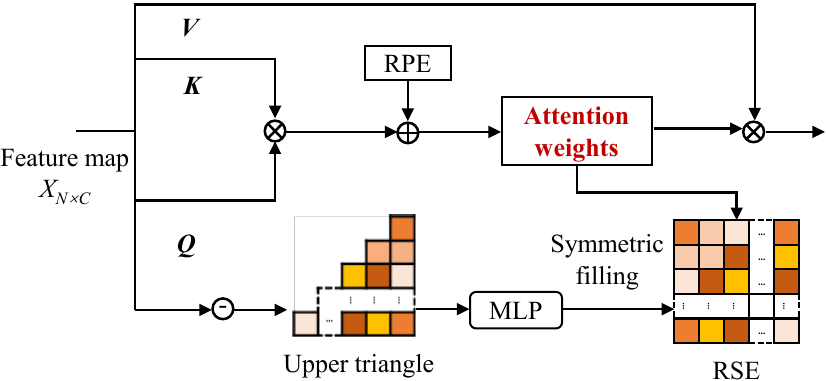}
   \caption{The framework of computing the Relative Semantic Embedding (i.e. RSE). We first obtain the relative semantic distance matrix by subtraction and convert it to the RSE with MLP. We use only the upper triangle of the distance matrix to ensure symmetry.}
   \label{fig:rse}
   \vspace{-0.3cm}
\end{figure}

\begin{table*}[t]
\centering
\caption{Comparison with state-of-the-art methods and baselines* on the PKU-DAVIS-SOD dataset~\cite{li2023sodformer}.}
\label{tab:pkudavissod}
\setlength{\tabcolsep}{2.48mm}{
\begin{tabular}{c|ccccc} 
\toprule
\textbf{Modality}            & \textbf{Method}                     & \textbf{Backbone}      & \textbf{Temporal} & \textbf{mAP$_{50}$} & \textbf{Runtime (ms)}  \\ \hline
\multirow{4}{*}{RGB}         & Faster R-CNN~\cite{ren2015faster}   & ResNet50+FPN           & No                & 0.443            & 11.6                   \\
                             & LSTM-SSD~\cite{liu2018mobile}       & Bottleneck-LSTM        & Yes               & 0.456            & 22.4                   \\
                             & Swin Transformer~\cite{liu2021swin} & Swin-T                 & No                & 0.475            & 14.6                   \\
                             & \textbf{SEST*}                      & Swin-T                 & No                & \textbf{0.486}   & 21.2                   \\ 
\hline
\multirow{2}{*}{Event}       & ASTMNet~\cite{li2022asynchronous}   & Rec-Conv-SSD           & Yes               & 0.291            & 21.3                   \\
                             & \textbf{Spiking Swin Transformer*}  & Swin QKFormer          & Yes               & \textbf{0.353}   & 28.9                   \\ 
\hline
\multirow{3}{*}{RGB + Event} & HDI-Former*                         & Swin-T + Swin QKFormer & Yes               & 0.501            & 47.6                   \\
                             & SODFormer~\cite{li2023sodformer}    & Deformable DETR        & Yes               & 0.504            & 39.7                   \\
                             & \textbf{HDI-Former}                 & Swin-T + Swin QKFormer & Yes               & \textbf{0.516}   & 47.9                   \\
\bottomrule
\end{tabular}}
\vspace{-0.3cm}
\end{table*}

\subsection{Framework of Relative Semantic Embedding}
For a more intuitive clarification of the proposed Relative Semantic Embedding (i.e., RSE), we show its computational procedure in Fig.~\ref{fig:rse}. The upper part of Fig.~\ref{fig:rse} indicates the computation of the original self-attention in Swin Transformer. On the bottom part, we calculate the relative semantic distance between any two tokens $a\in\mathbb{R}^{N\times N\times C}$ by subtraction as formulated in Eq.~\ref{eqn:6}. We denote this subtraction operation by the notation $\ominus$ in Fig.~\ref{fig:rse}. After that, we convert $a$ to RSE through an MLP with the output channel set to 1. Remarkably, considering that there are only two kinds of semantic relations between tokens in the object detection task: belonging or not belonging to the same object, the RSE here is supposed to be symmetric. Therefore, only the upper triangular part is fed into the MLP to ensure symmetry, which can be summarized as:
\begin{equation}
\mathcal{S}_{ij} = \left\{
   \begin{array}{l} 
   \bm{W}_{m}(\bm{W}_m'a_{ij} + \bm{B}_m') + \bm{B}_m,\quad i\leq j \\
   \\
   \mathcal{S}_{ji},\quad i > j
\end{array}.
\right.
\end{equation}

\section{More Details on Experimental Setting}

\subsection{Gen1 Dataset}
The Gen1 dataset is captured using a PROPHESEE GEN1 sensor with a resolution of 304×240 pixels. It consists of 2358 sequences, each containing recordings of open roads and various driving scenes for 60 seconds, encompassing urban environments, highways, suburban areas, and picturesque countryside landscapes. The object labels were meticulously assigned through a manual process and are available for two classes present: pedestrians and cars. The number of bounding boxes for pedestrians and cars is 228k and 27k, respectively.

\subsection{More Implementation Details}
\label{app:imp_detail}
Here we elaborate on some implementation details that are not covered in the Sec.~\ref{exp:set}. First, the spiking neuron adopted in our HDI-Former is LIFNode. In all our experiments, the reset value of LIF neurons $V_{reset}$ is set to 0, and the membrane time constant $\tau$ to 2. We set the threshold $V_{th}$ for the neurons following the output of self-attention to 0.5 and the rest to 1. The surrogate function engaged is the inverse trigonometric function (i.e., arctan function). The scalars attached to RPE, RSE, and dynamic interaction shown in Eq.~\ref{eqn:1}-\ref{eqn:5} (i.e., $\lambda_i, i=1,2,3,4)$ are set to $\lambda_1=\lambda_2=1,\lambda_3=0.3,\lambda_4=0.2$, which are decided by the experiments in Sec.~\ref{exp:lambda}. Owning to a trade-off between speed and accuracy, we simply use the attention kernel function described in Eq.~\ref{eqn:3} as $\bm{\mathcal{F}}$ in Sec.~\ref{met:dif}. Since QKA is equivalent to SSA with a diagonal attentional weight matrix, it benefits little from dynamic interaction. Therefore, the layers we conduct dynamic interaction start at Stage 3 in Fig.~\ref{fig:framework}. We use a learning rate scheduler that decreases the learning rate by 0.1 at epochs 27 and 33.

\begin{figure*}[t]
    \centering
    \centerline{\includegraphics[width=\linewidth]{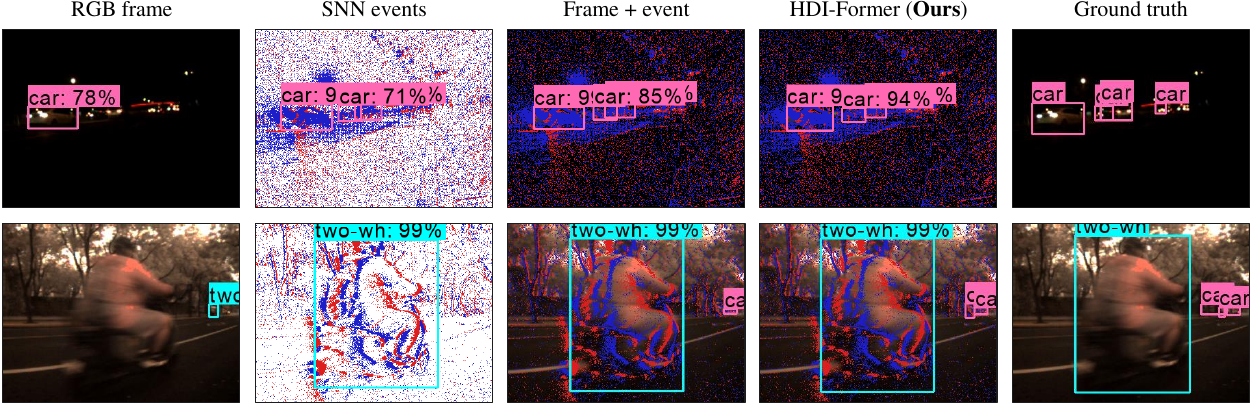}}
    \caption{Representative examples demonstrate that our HDI-Former effectively detects objects by integrating events and frames in challenging scenarios on the PKU-DAVIS-SOD dataset~\cite{li2023sodformer}.}
    \label{fig:pku_instance}
    \vspace{-0.18cm}
\end{figure*}

\section{More Experiments}
\subsection{Object Detection on PKU-DAVIS-SOD Dataset}
To verify the effectiveness and generality of our HDI-Former, we further conduct an extensive experiment on the PKU-DAVIS-SOD dataset~\cite{li2023sodformer}. As Table~\ref{tab:pkudavissod} shows, our HDI-Former consistently outperforms the state-of-the-art methods and the baselines* on the PKU-DAVIS-SOD dataset. Note that the mAP$_{50}$ improvement of 1.2\% compared to SODFormer~\cite{li2023sodformer} is even more significant than the 0.9\% improvement on the DSEC-Detection dataset. We attribute this to that the PKU-DAVIS-SOD dataset contains more challenging scenarios, and our HDI-Former shows a better ability to extract valid information from degraded modality with the help of dynamic interaction. We further present some instances under challenging scenarios (i.e., low light and high-speed motion blur) in Fig.~\ref{fig:pku_instance} to verify the effectiveness of integrating events and frames and the dynamic interaction mechanism in HDI-Former.

\begin{table}[t]
\centering
\caption{The influence of the scalars (i.e. $\lambda_1$-$\lambda_4$) involved in self-attention on the DSEC-Detection dataset~\cite{Gehrig24nature}.}
\label{tab:lambda}
\setlength{\tabcolsep}{3.97mm}{
\begin{tabular}{cccccc} 
\toprule
$\bm{\lambda_1}$ & $\bm{\lambda_2}$      & $\bm{\lambda_3}$ & $\bm{\lambda_4}$ & \textbf{mAP}   & \textbf{mAP$_{50}$}  \\ 
\hline
\multicolumn{2}{c}{\multirow{3}{*}{1}}   & 0.3              & 0.2              & \textbf{0.291} & \textbf{0.522}  \\
\multicolumn{2}{c}{}                     & 0.3              & 0.5              & 0.286          & 0.514           \\
\multicolumn{2}{c}{}                     & 0.5              & 0.2              & 0.289          & 0.518           \\
\bottomrule
\end{tabular}}
\vspace{-0.3cm}
\end{table}

\subsection{More Ablation Studies}
\label{exp:lambda}

\textbf{Influence of Scalars for RPE and RSE.}
To explore the influence of the hyperparameters in self-attention (i.e., $\lambda_1$-$\lambda_4$) and choose the best hyperparameter setting, we compare the performance of HDI-Former with varying $\lambda_i$ settings in Table~\ref{tab:lambda}. Particularly, we first follow the setting in \cite{liu2021swin} to set $\lambda_1$ and $\lambda_2$ to 1. For $\lambda_3$ and $\lambda_4$ which are involved in the dynamic interaction mechanism, we modify their settings to identify the best choice. From Table~\ref{tab:lambda}, we can observe that the performance is sub-optimal when $\lambda_4 > \lambda_3$. Moreover, the comparison between the first and the third rows in Table~\ref{tab:lambda} indicates that the performance is not especially sensitive to $\lambda_3$. Statistically, we conclude that relatively small $\lambda_3$ and $\lambda_4$ are sufficient to achieve our objective of cross-modality information interaction.

\begin{table}[t]
\centering
\caption{The influence of different attention kernel functions in dynamic interaction on the DSEC-Detection dataset~\cite{Gehrig24nature}.}
\label{exp:kernel}
\setlength{\tabcolsep}{4.03mm}{
\begin{tabular}{cccc} 
\toprule
\textbf{Method} & \textbf{mAP} & \textbf{mAP$_{50}$} & \textbf{Runtime (ms)}  \\ 
\hline
Baseline        & 0.291        & 0.522            & 50.3                   \\
Conv2D          & 0.294        & 0.529            & 61.0                   \\
MLP             & 0.294        & 0.531            & 73.7                   \\
\bottomrule
\end{tabular}}
\vspace{-0.3cm}
\end{table}

\noindent\textbf{Influence of Attention Kernel Functions.} 
The core of the dynamic interaction mechanism lies in the fact that the sharing of attention weights across modalities contributes to feature extraction. To achieve this, the attention kernel function (see Sec.~\ref{met:dif}) works similarly to the kernel function, projecting the attention weight of one modality into the attention weight space of the other modality since different modalities may have different attention weight spaces regarding head dimension as the channel. Akin to the kernel function, our attention kernel function is a set of functions. To explore the influence of attention kernel functions on the dynamic interaction mechanism, we compare the Baseline in Eq.~\ref{eqn:3} with more sophisticated functions like Conv2D or MLP in Table~\ref{exp:kernel}. As we can see, the HDI-Former achieves better performance with more powerful attention kernel functions. However, we state that the mAP improvements are marginal (i.e., 0.3\%) while increasing inference time. Therefore, our implementation in Eq.~\ref{eqn:3} reaches a good trade-off between performance and complexity.

\begin{figure}[t]
    \centering
    \centerline{\includegraphics[width=\linewidth]{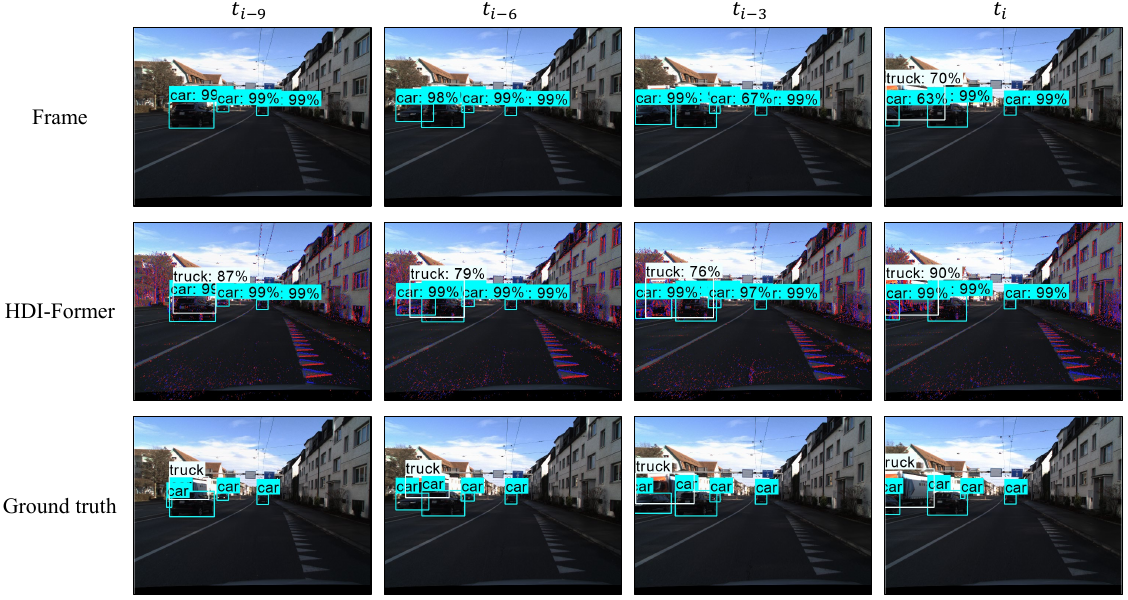}}
    \caption{Representative visualization results in continuous sequences on the DSEC-Detection dataset~\cite{Gehrig24nature}. The three rows are the frame-based SEST, our HDI-Former, and the ground truth.}
    \label{fig:temporal}
    \vspace{-0.3cm}
\end{figure}

\subsection{More Scalability Tests}

\textbf{Temporal Representation Analysis}
Since event streams contain rich temporal cues, HDI-Former is capable of utilizing the temporal cues to boost detection performance. To validate this, we conduct additional experiments on the DSEC-Detection dataset and provide some comparative visualization results of the HDI-Former and the feed-forward baseline (i.e., SEST) about whether they utilize rich temporal cues or not (see Fig.~\ref{fig:temporal}). Obviously, The baseline that utilizes single RGB frames for feed-forward object detection encounters challenges in scenarios involving occluded objects, as illustrated in the first row in which the occluded truck and the second car from the left are difficult to detect. However, our proposed HDI-Former effectively addresses this challenge by harnessing rich temporal information from continuous event streams, and thus successfully detects both objects with high confidence. Moreover, in contrast to existing models that utilize temporal cues, our approach generates temporal representation by processing event streams using SNN, reducing energy consumption.

\begin{table}[t]
\centering
\caption{Comparison of the inference time of the Baseline and the HDI-Former on the DSEC-Detection dataset~\cite{Gehrig24nature}.}
\footnotesize
\label{tab:runtime}
\setlength{\tabcolsep}{2.65 mm}{
\begin{tabular}{ccccc}
\toprule
\multirow{2}{*}{\textbf{Method}} & \multirow{2}{*}{\textbf{RSE}} & \multirow{2}{*}{\textbf{Events}} & \multirow{2}{*}{\textbf{\makecell[c]{Dynamic \\ Interaction}}} & \multirow{2}{*}{\textbf{Runtime(ms)}} \\ &  &  &  & \\
\hline
Baseline &  &  &  & 16.2  \\
(a) & $\surd $  &  &  & 24.7  \\
(b) & $\surd $  & $\surd $  &  & 50.2  \\
HDI-Former & $\surd $  & $\surd $  & $\surd $  & 50.3 \\ \bottomrule
\end{tabular}}
\end{table}

\noindent\textbf{Inference Time Analysis}
Following the same experimental setup as in Sec.~\ref{exp:set}, we tested the inference time of the baselines and our model on an NVIDIA A100 GPU. Unfortunately, the computation architecture of most existing computing hardware is not designed for SNN, and the A100 GPU is no exception. On one hand, the inference speed of SNN on such hardware is vastly reduced; on another hand, the two branches of our HDI-Former can be executed simultaneously, while the two branches can only be executed sequentially under the existing resources, which further reduces the inference frequency. Since our HDI-Former merely increases addition operations taking advantage of SNN and few multiplications, it is without doubt that the inference speed of our HDI-Former will increase significantly on neuromorphic computation hardware.

\begin{figure}[t]
    \centering
    \centerline{\includegraphics[width=\linewidth]{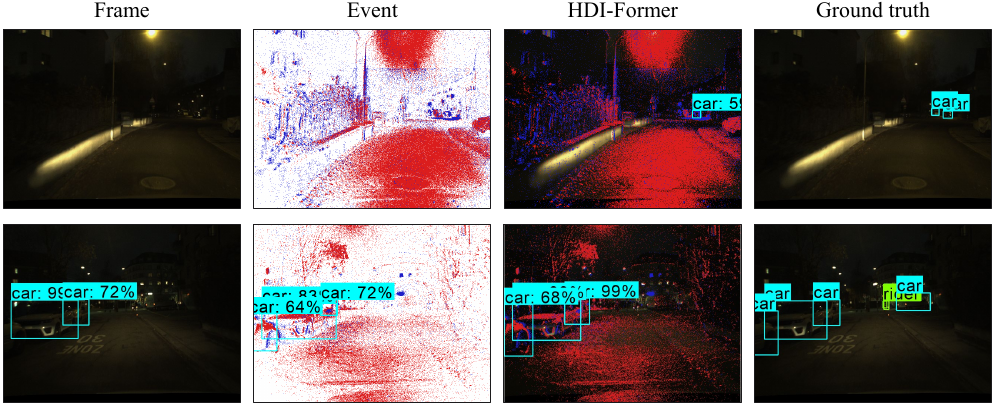}}
    \caption{Representative failure cases of our HDI-Former on the DSEC-Detection dataset~\cite{Gehrig24nature}.}
    \label{fig:failure}
    \vspace{-0.5cm}
\end{figure}

\noindent\textbf{Failure Case Analysis}
While our HDI-Former demonstrates promising performance even under challenging conditions, certain failure cases remain unresolved. In fact, some scenarios cannot be fully covered by the complementary properties of frames and events. This stems from the limited dynamic range of conventional frame-based cameras, resulting in poor image quality in such extreme lighting conditions. Conversely, event cameras excel at capturing moving objects but hardly output events for static or slow-motion objects. Therefore, some scenes pose a detection challenge to the current object detection paradigm, such as static objects or small objects in extreme lighting conditions. As illustrated in Fig.~\ref{fig:failure}, the four columns from left to right represent two unimodal baselines, our HDI-Former, and ground truth, respectively. The first row displays the difficulty of detection in the presence of small objects under low light even with our HDI-Former. In the second row, our HDI-Former fails to figure out the static car under low light on the right. Addressing these challenges remains an important area for future research, as they extend beyond the capabilities of our HDI-Former model.

\end{document}